\documentclass[letterpaper, 10 pt, conference]{ieeeconf}

\usepackage[T1]{fontenc}    % use 8-bit T1 fonts
\usepackage[utf8]{inputenc} % allow utf-8 input

\usepackage{url}            % simple URL typesetting
\usepackage{booktabs}       % professional-quality tables
\usepackage{amsfonts}       % blackboard math symbols

\usepackage[dvipsnames]{xcolor} % coloring of results in tables
\usepackage[svgnames]{xcolor}

\usepackage{amsmath} % assumes amsmath package installed

\usepackage{amsthm}
\usepackage{amssymb} % assumes amsmath package installed
\usepackage{mathtools}
\usepackage{mathrsfs}
\usepackage{mathdots}
\usepackage{bm}
\usepackage[linesnumbered,ruled,vlined]{algorithm2e}
\usepackage{graphics} % for pdf, bitmapped graphics files
\usepackage{svg}
\usepackage{tikz}
\usetikzlibrary{shapes, shapes.geometric, arrows, calc}
\usetikzlibrary{bayesnet}
\usetikzlibrary{fadings}
\usetikzlibrary{patterns}
\usepackage{yhmath}
\usepackage{cancel}
\usepackage{gensymb}

\usepackage{pgfkeys}
\usepackage{standalone}
\usepackage{fontawesome5} % For the solid star icon

\usepackage{subcaption}
\usepackage{cite}
\usepackage{graphicx}
\usepackage{multirow}
\usepackage{threeparttable}
\usepackage{adjustbox}
\usepackage{aligned-overset}
\usepackage{dblfloatfix}

\usepackage{float}
\usepackage{verbatim}
\let\labelindent\relax
\usepackage[shortlabels]{enumitem}
\usepackage{array}

\usepackage{hyperref} 
\hypersetup{
 colorlinks = true,
 linkcolor=black,
 urlcolor=Blue}
\definecolor{revise}{rgb}{0.54, 0.17, 0.89} % Blue violet
\IEEEoverridecommandlockouts                              % This command is only needed if 
\title{Learning to Throw-Flip}
\author{Yang Liu \quad Bruno Da Costa \quad Aude Billard
\thanks{This work was supported by the H2020 EU project DARKO under Grant Agreement No. 101017274.}
\thanks{The authors are with the Learning Algorithms and Systems Laboratory (LASA), EPFL, Switzerland. 
Corresponding author: Yang Liu 
({\tt\footnotesize yangliudh@gmail.com}).
}% <-this % stops a space
}
\begin{document}
\maketitle
\begin{abstract}
Dynamic manipulation, such as robot tossing or throwing objects, has recently gained attention as a novel paradigm to speed up logistic operations. However, the focus has predominantly been on the object's landing location, irrespective of its final orientation. In this work, we present a method enabling a robot to accurately ``throw-flip'' objects to a desired landing pose (position and orientation). Conventionally, objects thrown by revolute robots suffer from parasitic rotation, resulting in highly restricted and uncontrollable landing poses. Our approach is based on two key design choices: first, leveraging the impulse-momentum principle, we design a family of throwing motions that effectively decouple the parasitic rotation, significantly expanding the feasible set of landing poses. Second, we combine a physics-based model of free flight with regression-based learning methods to account for unmodeled effects. Real robot experiments demonstrate that our framework can learn to throw-flip objects to a pose target within $(\pm 5 \text{ cm}, \pm45 \text{ degree})$ threshold in dozens of trials. Thanks to data assimilation, incorporating projectile dynamics reduces sample complexity by an average of 40\% when throw-flipping to unseen poses compared to end-to-end learning methods. Additionally, we show that past knowledge on in-hand object spinning can be effectively reused, accelerating learning by 70\% when throwing a new object with a Center of Mass (CoM) shift. A video summarizing the proposed method and the hardware experiments is available at \href{https://youtu.be/txYc9b1oflU}{https://youtu.be/txYc9b1oflU}.

\end{abstract}
\setlength{\textfloatsep}{8pt} % Default ~20pt, reduce to tighten spacing
\section{Introduction}
Logistic operations, such as unpacking and sorting, often involve human operators throwing objects for efficiency. In contrast, robotic systems still rely on slower, quasi-static pick-and-place motions. Recent advances in robot throwing~\cite{zeng2020tossingbot, bombile2023bimanual, zermane2024planning} have introduced a new paradigm for fast and agile object transport, reducing both task cycle time and energy consumption. However, attention has so far been given solely to ensuring that the object lands precisely in a given location. However, precise control over the final orientation of objects when thrown remains underexplored, yet it is crucial for logistics operations, for instance, to enable accurate barcode scanning, maximize space utilization, and improve ergonomic access for human operators or other robots.

In this work, we explore controlled landing poses in the challenging task of throw-flipping a bar into a narrow box with a desired orientation (referred to as "throw-flip" hereafter for brevity). Throw-flipping is arguably one of the most complex dynamic manipulation tasks due to two key challenges:\\
\noindent \textbf{Coupled displacement and rotation:} For manipulators with revolute joints, increasing the linear velocity of the end-effector (EEF) to reach farther targets inherently increases its angular velocity, introducing a parasitic effect of exaggerated rotation. Thus, it is essential to design robot throwing actions that can independently steer both landing position and orientation.\\
\noindent \textbf{Intricate release dynamics:} The in-hand sliding and spinning that occur during the transient phase (approximately 50 ms) of vanishing gripper normal force are difficult to model. Additionally, some physical parameters, such as friction and deformations induced by a tight hold, are difficult to measure accurately.
% \begin{itemize}
%     \item \textbf{Coupled displacement and rotation:} For manipulators with revolute joints, increasing the linear velocity of the end-effector (EEF) to reach farther targets inherently increases its angular velocity, introducing a parasitic effect of exaggerated rotation. Thus, it is essential to design robot throwing actions that can independently steer both landing position and orientation.
%     \item \textbf{Intricate release dynamics:} The in-hand sliding and spinning that occur during the transient phase (approximately 50 ms) of vanishing gripper normal force are difficult to model. Additionally, some physical parameters, such as friction and deformations induced by a tight hold, are difficult to measure accurately.
%     % \item \textbf{Environmental factors:} While in indoor and for short-range / low-velocity throws, air drag can be neglected and the object's free-flying motion can be effectively described using the projectile motion of its center of mass (gravity-only) and the constant angular velocity of its rotation, this may not hold in all conditions. 
%     %An ideal flipping learning system should utilize this physical knowledge.
% \end{itemize}
\begin{figure}[t!]
    \centering
    \includegraphics[width=0.48\textwidth]{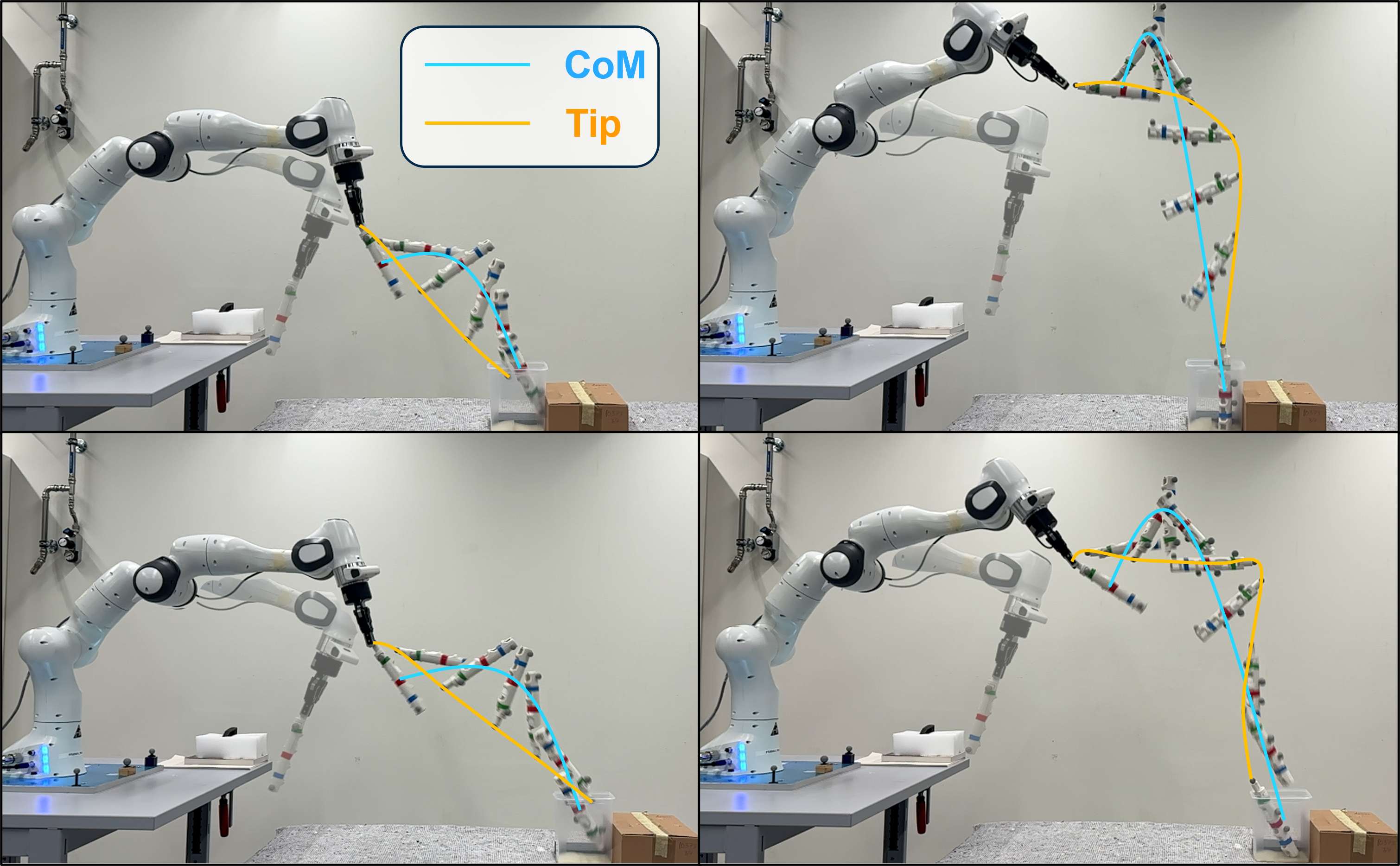}
    \caption{\small Robot throw-flips the bar with 4 different landing poses.}
    \label{fig1}
\end{figure}

Facing these challenges, we propose a framework that combines learning with physics, tailored to the throw-flipping task. Our contributions include,

\begin{itemize}
    \item \textbf{System design with impulse-momentum principle} to decouple displacement and rotation, resulting in drastically enlarged feasible landing poses.
    \item \textbf{Learning with data assimilation} 
    integrates empirical data with flying dynamics, reducing sample complexity by 40\% compared to pure data-driven learning.
    \item \textbf{Transfer learning to throw-flip a new object: } Reusing past data on in-hand object spinning when throw-flipping a new object under a Center of Mass (CoM) shift reduces sample complexity by 70\%, compared to CoM-shift-agnostic methods.
\end{itemize}

\section{Related work}
Over the past decades, there has been tremendous progress on robot throwing~\cite{aboaf1987task, schaal1993open, schaal1994robot, mason1993dynamic, lynch1999dynamic, senoo2008high, zhang2012sampling, pekarovskiy2013optimal,zeng2020tossingbot, monastirsky2022learning, 
liu2022solution,
bombile2023bimanual,
kasaei2023throwing,
pasala2024identification,  chen2024tossnet, liu2024tube, zermane2024planning, werner2024dynamic, ma2025learning}. Notably, much work has focused on throwing objects to desired \emph{landing locations}, with advancements in expanding operating conditions, e.g. throwing with dexterous postures~\cite{zhang2012sampling, liu2022solution, bombile2022dual, zermane2024planning} instead of planar throwing; throwing a variety of objects~\cite{zeng2020tossingbot, monastirsky2022learning, liu2024tube} rather than specific ones. However, robot throwing with desired \emph{landing poses (position and orientation)} has received much less attention, with only a few notable exceptions. Here, we review these in detail.\\
\textit{\textbf{Non-prehensile throwing with desired pose: }} In non-prehensile throwing, objects are held immobile via inertial forces during acceleration and are released instantaneously by maximizing deceleration of the hand to minimize the effects of friction during release, resulting in accurate throws. Most early works on robot throwing considered non-prehensile throwing~\cite{aboaf1987task, schaal1993open, schaal1994robot,mason1993dynamic, lynch1999dynamic}. In the challenging ``devil stick'' juggling task~\cite{schaal1993open, schaal1994robot}, a stick is manipulated to fly and spin between two hand sticks by hitting it back and forth. Memory-based learning was employed to refine the control of the stick's orientation, which is crucial for maintaining cyclic hits with symmetric rotation. However, the position of the center of mass is less critical and may exhibit significant error~\cite{devilsticking2016atkeson}. A fast learning method, Locally Weighted Regression, was used to construct a forward model, which was then inverted via pseudo-inversion for iterative command refinement. However, this approach does not guarantee the satisfaction of physical constraints or the generation of feasible commands. %In this task, as shown in the video~\cite{devilsticking2016atkeson}, devil sticking can be achieved without precise control over the throwing displacement.
\cite{lynch1999dynamic} applied sequential quadratic programming (SQP) for motion generation to throw a block with a flat pad to achieve a desired landing pose. More recently,~\cite{pekarovskiy2013optimal} adapted a similar setup in~\cite{lynch1999dynamic} to achieve high accuracy throwing by using the set of valid release states that result in successful throws. However, in these setups, the instantaneous release from the ``palm'' inherently binds the object's feasible free-flying motion to the motion capabilities of the end-effector. Consequently, for typical robot manipulators with revolute joints, the tight coupling between linear and angular motion at the end-effector severely restricts the range of achievable landing poses for the thrown object.\\
\textit{\textbf{Prehensile throwing with desired pose: }} Prehensile throwing, where the object is firmly grasped to prevent slippage during acceleration, poses additional challenges to achieve accurate throws due to release uncertainties arising from dynamic friction and deformations. Using a large amount of real throwing data (on the scale of thousands) to implicitly encapsulate all sources of uncertainty, including transient friction, TossNet~\cite{chen2024tossnet} learns an autoregressive model to predict object landing poses from joint motion and wrist force/torque sensor measurements before release. The trained model is then used to determine robot motions for accurate throws through a bisection method, including ``pitching'' bottles to desired poses. However, as demonstrated in the work, the obtained ``pitching'' motion only involves linear displacement without rotation (either for the end-effector or the flying object), implying that the landing orientation can only align with the end-effector's orientation at the moment of detach. Notably, TossNet also evaluated the accuracy of a benchmark physics model (rigid-body and projectile dynamics, neglecting frictional interactions), which showed significantly larger errors compared to the learned model ($\sim$3 cm vs. $\sim$15 cm error for throws with $\sim$50 cm horizontal displacement).

\noindent Compared to the literature that suffers from restricted landing poses, we aim to throw-flip objects to precise landing poses within a large and dense set of reachable outcomes. In addition, facing the dilemma between the inaccuracies of model-based planning and the high sample complexities of end-to-end learning, we aim to strike a balance by designing a learning system that seamlessly assimilates data with physical knowledge to accelerate learning.
\section{Method}
% A schematic of the throw-flip variables is illustrated in Fig.~\ref{fig:notation}. The robot base frame $\mathcal{A}$ serves as world frame. Hand frame $\mathcal{H}$ is located at the center of the two fingers. Object frame $\mathcal{O}$ is located at its center of mass (CoM). In the absence of external disturbance, the throw-flip motion is planar.
% In the plane of motion, the coordinates of the hand frame w.r.t. the world frame reduce to three parameters: $\mathbf{q}^h = [x^h, z^h, \theta^h]^\top \in \mathbb{R}^3$, where $x^h$ and  $z^h$ are the coordinates on the horizontal and vertical axes, and $\theta^h$ the orientation of the object at release time relative to the vertical axis. Hand twist at the center of the two fingers is expressed in the world frame $\mathcal{A}$: $\mathbf{v}^h = [v^h_{x}, v^h_{z}, \omega^h]^\top \in \mathbb{R}^3$.
\subsection{Modeling throw-flipping}
\label{subsec:modeling}
\begin{figure}[t!]
    \centering
\includegraphics[width=\linewidth, trim=0cm 1cm 0cm 0.8cm, clip]{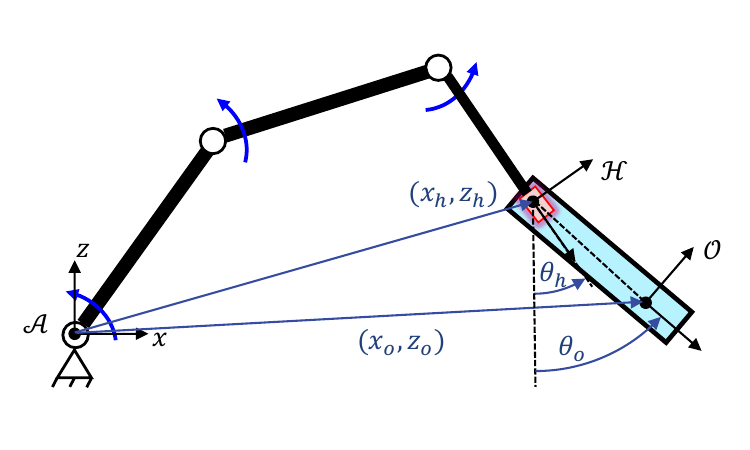}
    \caption{\small Major notations for flipping.}
    \label{fig:notation}
\end{figure}

A schematic of the throw-flip variables is illustrated in Fig.~\ref{fig:notation}. The robot base frame $\mathcal{A}$ serves as world frame. Hand frame $\mathcal{H}$ is located at the center of the two fingers. Object frame $\mathcal{O}$ is located at its center of mass (CoM). In the absence of external disturbance, the throw-flip motion is planar.
In the plane of motion, the coordinates of the hand frame w.r.t. the world frame reduce to three parameters: $\mathbf{q}^h = [x^h, z^h, \theta^h]^\top \in \mathbb{R}^3$, where $x^h$ and  $z^h$ are the coordinates on the horizontal and vertical axes, and $\theta^h$ the orientation of the object at release time relative to the vertical axis. Hand twist at the center of the two fingers is expressed in the world frame $\mathcal{A}$: $\mathbf{v}^h = [v^h_{x}, v^h_{z}, \omega^h]^\top \in \mathbb{R}^3$.

Denote the joint position and velocity of an N-DoF robot manipulator as $\mathbf{q} \in \mathbb{R}^N$, $\dot{\mathbf{q}} \in \mathbb{R}^N$, respectively. Associate the end-effector (EEF) states in the world frame using standard forward kinematics and differential forward kinematics relationships: 
\begin{align*}
\mathbf{q}^h &=\mathbf{f}_{kin}^h(\mathbf{q}): \mathbb{R}^N \rightarrow \mathbb{R}^3\\
\mathbf{v}^h &= \mathbf{J}^h(\mathbf{q}) \dot{\mathbf{q}}: \mathbb{R}^N\times \mathbb{R}^N \rightarrow \mathbb{R}^3
\end{align*}
The generalized coordinates of the object/bar are denoted by $\mathbf{q}^o = [x^o, z^o, \theta^o]^\top \in \mathbb{R}^3$. %, where $x^o$ is the horizontal position of its CoM relative to the robot base frame $\mathcal{A}$, $z^o$ is the vertical position of its CoM relative to the robot base frame $\mathcal{A}$, $\theta^o$ is the object's orientation relative to the vertical orientation. 
Note that $\theta^o$ is unwrapped from $[0, 2\pi]$ in order to differentiate different number of flips during its free flying. Likewise, object twist at CoM is expressed in the world frame $\mathcal{A}$ and is denoted by $\mathbf{v}^o = [v^o_{x}, v^o_{z}, \omega^o]^\top \in \mathbb{R}^3$.

For indoor throwing of solid objects, air drag can be neglected. The free-flying dynamics is only subject to gravitational force, resulting in the object acceleration $\mathbf{a}^o := [a^o_x, a^o_z, \alpha^o]^\top = [0, -G ,0]^\top$, where $G = 9.81 \text{ m}/\text{s}^2$ is the gravitational acceleration. For such projectile dynamics, the landing pose can be obtained in closed form given the release state $(\mathbf{q}^o, \mathbf{v}^o)$. The flying duration $t_{fly}$ can be calculated as $t^{fly} = \left(\dot{z}^o+\sqrt{\dot{z}^{o^2}+2G(z^o-z^{land})}\right)/G$.
% \begin{align*}
%     t^{fly} = \frac{\dot{z}^o+\sqrt{\dot{z}^{o^2}+2G(z^o-z^{land})}}{G}
% \end{align*}
% $t^{fly} = \frac{\dot{z}^o+\sqrt{\dot{z}^{o^2}+2G(z^o-z^{land})}}{G}$. 
Then, $x^{land} = x^o+v^o_x t^{fly}, \theta^{land} = \theta^o+\omega^o t^{fly}$. Without loss of generality, in this work, we assume that the landing height is always 0 in the robot frame and define the landing pose at height 0 as $\mathbf{q}^{land} = (x^{land}, \theta^{land})$. We define the projectile flying flowmap $g$, with $\mathbf{q}^{land} = g(\mathbf{q}^o, \mathbf{v}^0)$, to compute the landing pose at 0 height given release state.

Define the relative coordinates of object frame $\mathcal{O}$ w.r.t. hand frame $\mathcal{H}$, $\mathbf{q}^{r} = [x^{r}, z^{r}, \theta^r]^\top := \mathbf{q}^o - \mathbf{q}^h \in \mathbb{R}^3$. Then for the contact point $\mathcal{C}$ on the object specified by a relative vector $\mathbf{q}^{r}$, the contact point twist $\mathbf{v}^c$ and  object twist $\mathbf{v}^o$ are related through:
\begin{align}
\mathbf{v}^c = \begin{bmatrix}
v^c_x\\
v^c_z\\
w^c
\end{bmatrix}
= \begin{bmatrix}
v^o_x+w^o z^r\\
v^o_z-w^o x^r\\
w^o
\end{bmatrix}
\label{equation:vc}
\end{align}
% =
% \mathbf{G}(\mathbf{q}_r) \mathbf{v}^o
% with
% \begin{align}
%  \mathbf{G}(\mathbf{q}_r) =
% \begin{bmatrix}
% 1 & 0 & z^r \\
% 0 & 1 & -x^r \\
% 0 & 0 & 1
% \end{bmatrix},
% \end{align}

The relative velocity $\mathbf{v}^{r}$ is defined as the difference between the velocity of the \emph{contact point} $\mathbf{v}^c$ and the velocity of the hand $\mathbf{v}^h$, i.e. $\mathbf{v}^{r} := \mathbf{v}^c - \mathbf{v}^h$.

\subsection{Impulse-momentum-based control design}
\label{subsec:control-design}
\begin{figure}[h!]
    \centering
\includegraphics[width=\linewidth, trim=0cm 0cm 0cm 0.0cm]{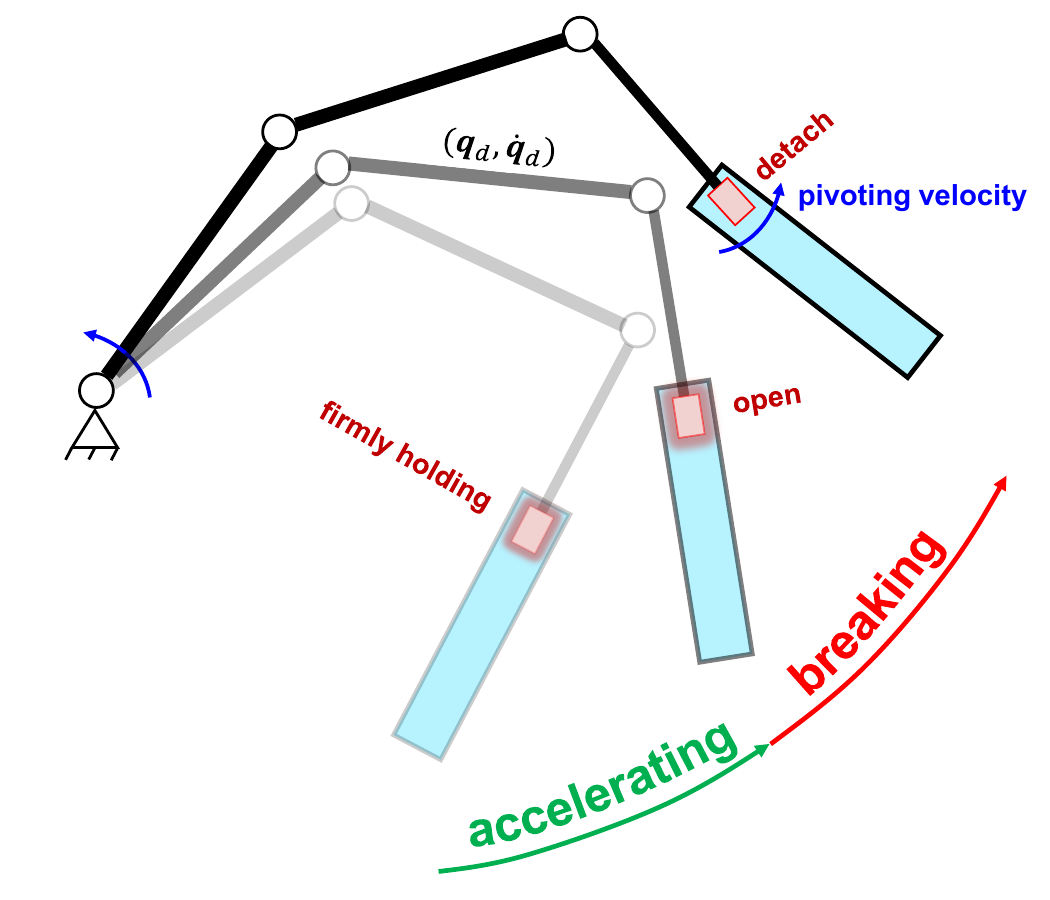}
    \caption{\small Schematic of flip motion. The robot firmly grasps the bar and accelerates it to a high-energy, high-velocity state, then enters a rapid decelerating-braking phase. During braking, the robot's gripper begins to open with decreasing normal force. Approximately 50 ms later, the normal force vanishes completely, and the bar enters free flight. Leveraging the impulse-momentum principle, the bar's rotation is accelerated by the pivoting velocity.}
    \label{fig:flip-motion}
\end{figure}

% Conventionally, a real robot's throwing motion with prehensile grasps is driven by a precomputed or learned throwing trajectory towards a desired end-effector state (pose and twist). Considering the inevitable transient window of gradually decreasing normal force in common parallel-draw grippers, prior works either learn to adjust the throwing trajectory to compensate for release delay implicitly through domain randomization~\cite{monastirsky2022learning} or synthesize a robust throwing trajectory against release uncertainty explicitly using convex optimization~\cite{liu2024tube}. Regardless of the exact throwing trajectory generation method, such trajectory-based throwing motions face a fundamental limitation in the flipping task for revolute robots: an enlarged end-effector (EEF) linear velocity is always accompanied by increased angular velocity, making it difficult for the robot to independently control the landing position and orientation. Furthermore, given that the length of the kinematic chain in common manipulators (80 cm - 120 cm) is significantly longer than the thrown bar (24 cm in our case), it is impossible for the robot to flip the bar into a nearby cup.
For manipulators with revolute joints, the coupling between the linear motion and the angular motion of the end-effectpr significantly limits the feasible space of the landing poses of thrown objects. To address this challenge, we propose leveraging the temporal hinge effect during the gripper-opening window, as illustrated in the schematic of Fig.~\ref{fig:flip-motion}. The temporal hinge can be demonstrated through a simple dropping experiment with our fingers: pinch-grasp a pen at one end, hold it horizontally, and gradually open the fingers—the pen will first pivot in-hand and then detach. During the transient release window, the finger-bar contact behaves like an underactuated joint, making the robot-bar system an (N+1)-DoF pendubot for an N-DoF manipulator. In this scenario, if the robot undergoes drastic deceleration during the temporal hinge window, the impulse-momentum principle allows part of the robot's angular momentum to be transmitted to the bar through the frictional interface, accelerating the object's angular velocity. The amount of hinge acceleration can be controlled by the magnitude of braking during this window. As a result, we obtain one additional steerable DoF, independent of the nominal throwing state.
\subsection{3-parameter family of flip motions}
\colorlet{MildRed}{Maroon}
\colorlet{MildGreen}{OliveGreen}
\colorlet{MildBlue}{NavyBlue}
As sketched in Section~\ref{subsec:control-design}, the two-phase flip motion family proceeds as follows:
\subsubsection{\textbf{Acceleration}} 
Given a desired nominal throwing state in joint space $(\mathbf{q}_d, \mathbf{\dot{q}}_d) \in \mathbb{R}^{2N}$, a dynamically feasible accelerating trajectory is generated to connect the pre-throw pose with the nominal state. To ensure dynamic feasibility, the trajectory is generated using PolyMPC~\cite{listov2020polympc}, a trajectory optimizer that enforces dynamic constraints (joint position, velocity, acceleration, and torque limits) while minimizing time, given the initial robot state and the nominal throwing state $(\mathbf{q}_d, \mathbf{\dot{q}}_d)$. This trajectory is then executed via a standard impedance controller. %The gripper opening command is given 50 ms before the end of the acceleration trajectory to account for the dwell time of the gripper.

Starting from a reference throwing state $(\mathbf{q}_{\text{ref}}, \mathbf{\dot{q}}_{\text{ref}}) \in \mathbb{R}^{2N}$, the nominal throwing state $(\mathbf{q}_d, \mathbf{\dot{q}}_d)$ is obtained by adjusting two scalar parameters, \textcolor{MildRed}{``pitch'' $\gamma$} and \textcolor{MildGreen}{``speed'' $s$}, as follows:
\begin{align*}
   \mathbf{q}_d &= \mathbf{q}_{\text{ref}} + \frac{\textcolor{MildRed}{\gamma}}{N}I_{N\times 1}, \quad\mathbf{\dot{q}}_d = \textcolor{MildGreen}{s}\mathbf{\dot{q}}_{\text{ref}} 
\end{align*}
Effectively, \textcolor{MildRed}{``pitch'' $\gamma$} and \textcolor{MildGreen}{``speed'' $s$} regulate the direction and the magnitude of the finger's linear velocity.
\subsubsection{\textbf{Brake}} After bypassing the end of the trajectory, the target position is set as $\mathbf{q}_d$ and the target velocity is set to $\mathbf{0}$ for the impedance controller. The magnitude of the brake is regulated by the \textcolor{MildBlue}{``damping'' $D$} parameter of the impedance controller.

In summary, the family of robot flip motions is indexed by (\textcolor{MildRed}{``pitch'' $\gamma$}, \textcolor{MildGreen}{``speed'' $s$}, \textcolor{MildBlue}{``damping'' $D$}). Hence, we define the flip command $\mathbf{u}:=(\gamma, s, D) \in \mathbb{R}^3$.

\subsection{Learning and Adaptation}
\begin{figure*}[th!]
\centering
\includegraphics[width=0.9\textwidth, trim=0mm 0mm 0mm 0mm, clip]{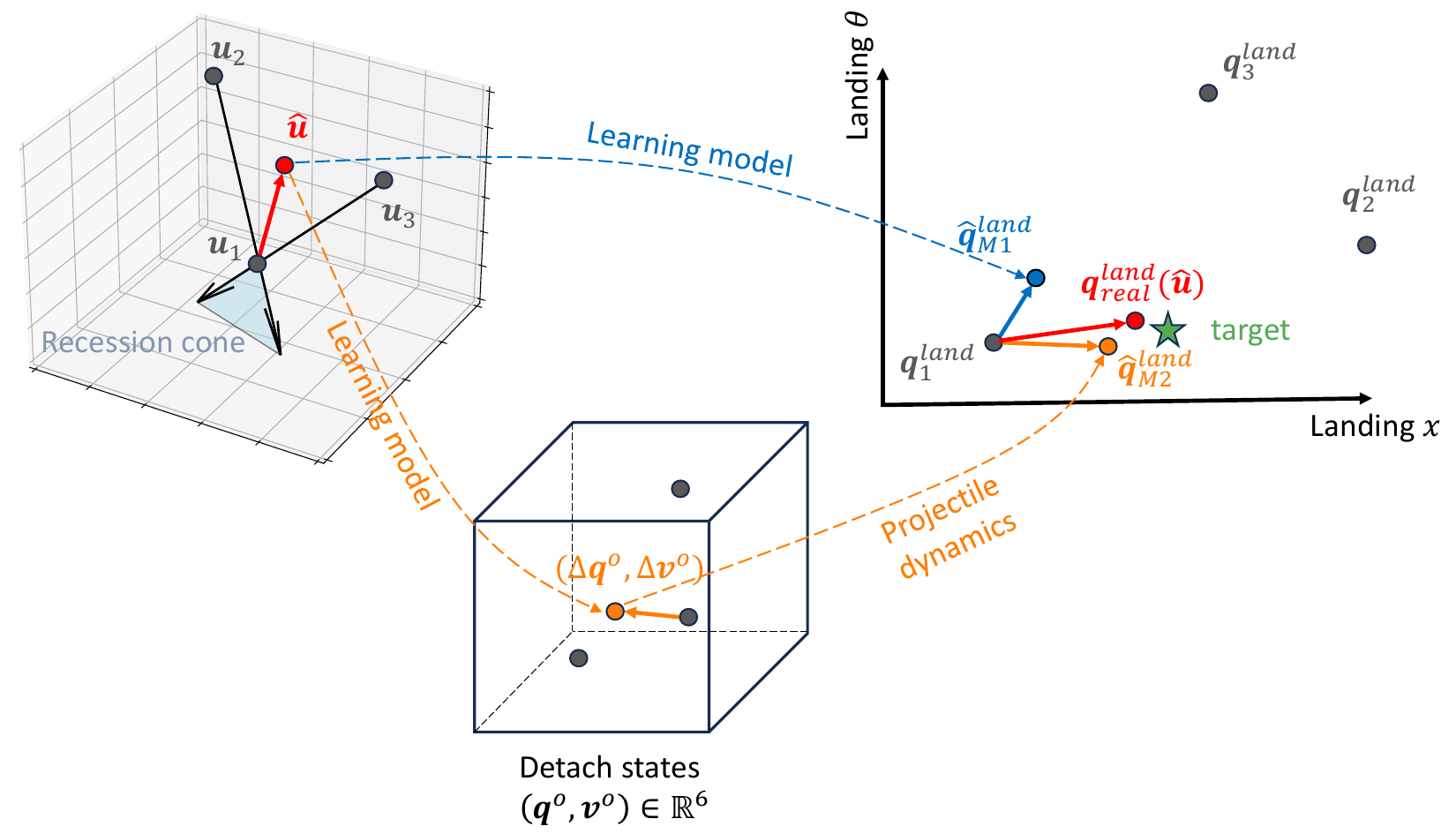}
%\includesvg[width=0.5\textwidth]{imgs/landing_x_angle_scatter}
\caption{\small Schematic of the two forward models to predict the landing pose of a new command \textcolor{red}{$\mathbf{\hat{u}}$}.}
\label{fig:forward-schematic}
\end{figure*}
% \begin{figure*}[ht!]
%     \centering
% \includegraphics[width=\linewidth, trim=0cm 1cm 0cm 1.0cm]{imgs/iterative_learning_diagram.pdf}
%     \caption{\small Diagram of the process of learning to flip.}
%     \label{fig:learning-diagram}
% \end{figure*}
We aim to solve the \emph{inverse problem} of finding appropriate control commands for desired throwing outcomes through iterative learning. The rationale behind the data-driven approach is the absence of accurate \emph{forward models} to describe the physical process of throw-flipping. This is not only due to the complex and intricate dynamic patch friction between the finger pad and the object, but also due to the lack of accurate robot dynamics models when operating in the highly dynamic regime of impulse-momentum braking. 

Our learning to throw-flip setup is as follows: Given an initial support dataset $\mathcal{X}_0=\{(\mathbf{u}, \mathbf{q}^h, \mathbf{q}^r, \mathbf{v}^h, \mathbf{v}^r, \mathbf{q}^o, \mathbf{v}^o, \mathbf{q}^{land})\}$ representing the initial observations of the physical process, the task is to iteratively refine control commands that progressively throw-flips the object closer to an unseen target landing pose as new data is gathered throughout the process. For a learning and adaptation system, two crucial design choices must be considered:
\begin{itemize}
    \item \textbf{Forward model: } Given the dataset of past experiences, how to build a forward model to \emph{predict} the throwing outcome of a new command? 
    \item \textbf{Iterative command adaptation: } How to effectively adapt the command based on the updated forward model? As an autonomous decision-maker, what actions should be taken when the process gets stuck?
\end{itemize}
Next, we describe our iterative learning design in detail.
\subsection{Forward Model}
Despite the absence of accurate release dynamics models and robot dynamics models, there is still rich geometrical and physical knowledge about the throw-fliping process at our disposal, as presented in Section~\ref{subsec:modeling}. Can this knowledge be utilized to accelerate learning? To answer this question, we design and compare the following two models, as illustrated in Fig.~\ref{fig:forward-schematic}: 
\begin{itemize} 
\item \textbf{Model 1: end-to-end} learns a locally linear map from the 3D command $\mathbf{u}$ to the 2D landing pose directly. 
\item \textbf{Model 2: projectile-based} learns a locally linear map from 3D command to the 6D object's detach state $(\mathbf{q}^o, \mathbf{v}^o)$ and then obtains the predicted landing pose using the projectile flying flowmap $g(\mathbf{q}^o, \mathbf{v}^o)$. Compared to \textbf{Model 1}, this model encodes physical knowledge of flying dynamics to assist learning.
\end{itemize}

\noindent Mathematically, we define the normalized error as,
\begin{align*}
    e(\mathbf{q^{land}, \mathbf{q}^{target}})= \left\|\left[\frac{x^{land}-x^{target}}{\epsilon_x}, \frac{\theta^{land}-\theta^{target}}{\epsilon_\theta}\right] \right\|_2
\end{align*} where $\epsilon_x, \epsilon_\theta$ are the target pose thresholds for position and orientation, respectively. From a dataset $\mathcal{X}$, we identify the three nearest neighbors based on the normalized error, denoted as \textcolor{blue}{$(\mathbf{u}_i, \mathbf{q}^o_i, \mathbf{v}^o_i, \mathbf{q}^{land}_i), i\in{1,2,3}$}, ranked by their normalized error relative to the target pose. Define the delta command as $\Delta \mathbf{u}(\alpha_1, \alpha_2) = \sum_{i=1,2}\alpha_i (\mathbf{u}_{i+1} - \mathbf{u}_1)$, where $(\alpha_1, \alpha_2) \in \mathbb{R}^2$ is the recession cone coordinate we are searching for. Then the two \emph{forward models} predict the landing poses of a new command $\mathbf{u} = \mathbf{u}_1+\Delta \mathbf{u}(\alpha_1, \alpha_2)$ as follows:
\begin{align}
    &\hat{\mathbf{q}}^{land}_{M1}(\alpha_1, \alpha_2) 
    &= \textcolor{blue}{\mathbf{q}^{land}_1} + \sum_{i={1,2}}\alpha_i (\textcolor{blue}{\mathbf{q}^{land}_{i+1}- \mathbf{q}^{land}_1})
    \label{eq:m1}
\end{align}
\begin{align}
\begin{split}
    &\hat{\mathbf{q}}^{land}_{M2}(\alpha_1, \alpha_2)  = g(\hat{\mathbf{q}}^o, \hat{\mathbf{v}}^o)\\
    &= g(\mathbf{q}_1^o+\sum_{i=1,2}\alpha_i(\textcolor{blue}{\mathbf{q}^{o}_{i+1}- \mathbf{q}^{o}_1}), \mathbf{v}_1^o+\sum_{i=1,2}\alpha_i(\textcolor{blue}{\mathbf{v}^{o}_{i+1}- \mathbf{v}^{o}_1}) )
\end{split}
    \label{eq:m2}
\end{align}

% \begin{figure}[h!]
%     \centering
% \includegraphics[width=\linewidth, trim=0cm 0cm 0cm 0.5cm]{imgs/scm_plot_2.pdf}
%     \caption{\small Structure of the forward model. The dashed arrows from the control parameters $(\gamma, s, D)$ to the hand/relative states $(\mathbf{q}^h, \mathbf{q}^r, \mathbf{v}^h, \mathbf{v}^r)$ indicate the absence of accurate physical models to describe the relationship, whereas the solid arrows indicate that accurate physical models are readily available.}
%     \label{fig:scm-plot}
% \end{figure}

\subsection{Model Inversion with On-Manifold Cone Search}
In essence, our iterative learning problem is a zeroth-order optimization/root-finding problem, where only sparse forward model evaluation is available. At each iteration, given the current best command (\textcolor{blue}{$\mathbf{u}_1$}) for throw-flipping to the desired target, we aim to determine the delta change of the current best command ($\Delta \mathbf{u}$) to throw closer. We propose a fast adaptation method using on-manifold exhaustive search in the local recession cone. At each iteration, the robot samples a dense mesh of candidate feasible commands parametrized by the recession cone coordinate $(\alpha_1, \alpha_2)$. It then selects the best candidate that minimizes the \emph{predicted error} to execute and adds the new observations to the dataset. Progressively, the algorithm refines the throwing command using more accurate forward models around the target in the accumulated dataset. The pseudocode describing the procedure is summarized in Algorithm 1.

% \begin{figure}[t!]
%     \centering
%     % \captionsetup{aboveskip=5pt, belowskip=5pt}
%     \includegraphics[width=0.4\textwidth, trim=5mm 25mm 5mm 10mm, clip]{imgs/recession_cone.pdf}
%     \caption{\small Illustration of the recession cone coordinate.}
%     \label{fig:recession_cone}
% \end{figure}
\begin{algorithm}[t!]
\caption{Iterative Learning with Cone Search}
\SetKwInput{KwInput}{Input} 
\SetKwInput{KwInit}{Initialize} 

\KwInput{Support dataset $\mathcal{X}$, Target pose $\mathbf{q}^{target}$, Max iterations $T$, Error thresholds $\epsilon_x, \epsilon_\theta$, $\#$Command trials $N$}

\For{$t = 1$ to $T$}{
    Identify three nearest neighbors in $\mathcal{X}$:  
    $(\mathbf{u}_i, \mathbf{q}^o_i, \mathbf{v}^o_i, \mathbf{q}^{land}_i), \quad i \in \{1,2,3\}$\;
    
    \ForEach{$(\alpha_1, \alpha_2)$ in mesh grid}{
        Compute $\hat{\mathbf{q}}^{land}$ using Eq.~\ref{eq:m1} or Eq.~\ref{eq:m2} \;
        Compute error $e(\hat{\mathbf{q}}^{land}, \mathbf{q}^{target})$\;
    }

    Find best command $\mathbf{u}^{t}(\alpha_1, \alpha_2)$ minimizing $e$
    % and update :
    % \[
    % \mathbf{u}^{t} = \mathbf{u}_1 + \sum_{i=1}^{2} \alpha_{i+1} (\mathbf{u}_i - \mathbf{u}_1)
    % \]
    Execute $\mathbf{u}^{t}$, observe $N$ trials, compute mean landing pose $\bar{\mathbf{q}}^{land}_t$
    % \[
    % \bar{\mathbf{q}}^{land}_t = \frac{1}{N} \sum_{n=1}^{N} \mathbf{q}^{land}_{t,n}
    % \]

    Update dataset: $
    \mathcal{X} \gets \mathcal{X} \cup (\mathbf{u}^{t}, \bar{\mathbf{q}}^{land}_{t})$

    \If{All $N$ trials land within thresholds}{
        \Return $\mathbf{u}^{t}$ (Success)\;
    }

    \If{learning stagnates (error does not improve)}{
        Select new neighbors to escape local minimum\;
    }
}
\Return best found $\mathbf{u}$\;
\end{algorithm}

\subsection{Transfer Learning under CoM shift}
%We investigate the extent to which our physics-based learning approach can ease transfer learning and consider the problen of throwing a new object with a different CoM. 
To throw a new object with a different and known CoM, instead of gathering a support set from scratch and then starting iterative learning, we propose a method to transfer experience from throwing other objects to accelerate the learning process for the new object.

Assuming that the in-hand sliding-spinning twist remains unchanged after a CoM shift $\Delta h$, according to Eq.~\ref{equation:vc}, we obtain the predicted landing pose of the new object by shifting the detach state $(\mathbf{q}^o, \mathbf{v}^o)$ as follows:
\begin{align*}    \hat{\mathbf{q}}^o(\Delta h) &=  \mathbf{q}^o + \Delta h[\sin \theta^o, -\cos \theta^o , 0 ]^\top\\
\hat{\mathbf{v}}^o(\Delta h) &=  \mathbf{v}^o + \omega^o\Delta h[\cos \theta^o, \sin \theta^o , 0 ]^\top
\end{align*}
\noindent Integrating $(\mathbf{q}^o, \mathbf{v}^o)$ with projectile dynamics yields the transferred support set $\hat{\mathcal{X}}$. The learning procedure, denoted as \textbf{Model 3: transfer learning}, is as follows,
\begin{itemize}
    \item \textbf{1st Iteration: } Select the closest command within the transferred support set $\hat{\mathcal{X}}$, denoted as $\mathbf{u}_1$. After executing $\mathbf{u}_1$, obtain data \textcolor{blue}{$(\mathbf{q}^o_1,\mathbf{v}^o_1, \mathbf{q}^{land}_1)$}.
    \item \textbf{2nd Iteration: }
    Select the 3 closest data entries within $\hat{\mathcal{X}}$, denoted as $\{(\hat{\mathbf{u}}_i, \hat{\mathbf{q}}^o_i,\hat{\mathbf{v}}^o_i, \hat{\mathbf{q}}^{land}_i)\}, i \in {1,2,3}$.% (Note that $\mathbf{u}_i == \hat{\mathbf{u}}_i$).
    Then we compute the best delta command $\Delta \mathbf{u}$, such that $\mathbf{u}_2 = \mathbf{u}_1 + \Delta\mathbf{u}$ yields closest landing pose $\hat{\mathbf{q}}^{land}_2$ predicted by the 3 selected data. In particular, let $\Delta \mathbf{u}(\alpha_1, \alpha_2) = \sum_{i=1,2}\alpha_i (\hat{\mathbf{u}}_{i+1}-\hat{\textbf{u}}_1)$, then
    \begin{align*}
    \begin{split}
    &\hat{\mathbf{q}}^{land}_2 = g(\hat{\mathbf{q}}^o_2, \hat{\mathbf{v}}^o_2)=\\
     &g(\textcolor{blue}{\mathbf{q}_1^o}+\sum_{i=1,2}\alpha_i(\textcolor{black}{\hat{\mathbf{q}}^{o}_{i+1}- \hat{\mathbf{q}}^{o}_1}), \textcolor{blue}{\mathbf{v}_1^o}+\sum_{i=1,2}\alpha_i(\textcolor{black}{\mathbf{\hat{v}}^{o}_{i+1}- \mathbf{\hat{v}}^{o}_1}) )
    \end{split}
    \end{align*}
    %= g(\mathbf{q}^o_1+\Delta \mathbf{q}^o(\Delta u),\mathbf{v}^o_1+\Delta \mathbf{v}^o (\Delta u)) $
    %. After executing $\mathbf{u}_1$, obtain landing pose $\mathbf{q}^{land}_2$;
    % In other words, the first iteration data $(\mathbf{q}^o_1,\mathbf{v}^o_1)$ serves as a ``calibration'' to adjust the prediction of $\hat{\mathbf{u}}_1$, and the delta dynamics from $\hat{\mathcal{X}}$ is used to estimate the delta change of intermediate variables $(\Delta \hat{\mathbf{q}}^o, \Delta \hat{\mathbf{v}}^o)$ given the delta change in the command $\Delta \mathbf{u}$.
    \item \textbf{3rd Iteration: }
    Select the 1st, 2nd and 4th closest data entries within $\hat{\mathcal{X}}$, compute the best delta command $\Delta \mathbf{u}$, such that $\mathbf{u}_3 = \mathbf{u}_2 + \Delta\mathbf{u}_2$ yields the closest predicted landing pose $\hat{\mathbf{q}}^{land}_3$, calculated in the same way as in the 2nd Iteration. $\hat{\mathbf{q}}^{land}_3$ is calculated the same way as in the 2nd iteration,
    \item \textbf{From the 4th Iteration Onward: } Start normal iterative learning using a new support set, consisting of the data obtained from the first three iterations.
\end{itemize}

% % \subsection{Optimization-based command adaptation}
% Since the dimension of the control command ($\mathbb{R}^3$) is larger than that of the landing pose ($\mathbb{R}^2$), there exists a ``Nullspace'' of valid commands that lead to the same target. By utilizing numerical optimization for redundancy resolution, the obtained control command can be explicitly constrained within the feasible set. Mathematically, the problem for command adaptation is formulated as follows:
% \begin{align*}
% \textbf{Problem \textit{Adaptation}}\\
% %\min_{\overrightarrow{A B}, q, \dot{r}, \dot{z}} \quad 0 \\
% \text { Find: } & \left\{ \textcolor{black}{\mathbf{u}} \in \mathbb{R}^3 \right\} \quad \\
% \text { subject to: } 
% % & \textcolor{black}{q(0)} = q_d, \quad \tag{1b}\\
% % & \textcolor{black}{\dot{q}(0)} = \dot{q}_d, \quad \tag{1c}\\
% & \|f^H(\mathbf{u}, \mathbf{\varphi}) - \mathbf{q}^{land}_d)\| \leq \epsilon, \quad \\
% & \mathbf{u} \in \mathcal{U}
% \label{problem-adapt}
% \end{align*}

% here, $\epsilon$ is a predefined threshold based on the size of the target cup, and $\mathcal{U}$ is the feasible command set, which can be determined by executing throwing motion without the object, plus a safety margin that accounts for the dynamic reaction from the payload during throwing. 

% \textcolor{blue}{TODO (Bruno): add the exact learning and adaptation algorithms}
\section{Experiment}
\subsection*{Hardware setup}
The throwing experiments are conducted using a 7-DOF Franka Emika Panda manipulator mounted with a Robotiq 2F-85 parallel gripper. The thrown object is a 3D-printed bar attached with reflective OptiTrack markers, enabling precise 6-DoF tracking of the object’s pose both during free flight and upon landing. Its mass distribution can be configured by shifting the payload position within the bar.

% \begin{figure}[h!]
% \centering\includegraphics[width=0.35\textwidth]{imgs/throw_bar.jpg}
%     \caption{\small The thrown object: a 3D-printed bar with configurable mass distribution attached with markers.}
% \label{fig:thorw-bar}
% \end{figure}

\subsection{Control design verification}
\label{subsec:control-verification}
\begin{figure}[h!]
\centering
\includegraphics[width=0.4\textwidth, trim=0mm 0mm 0mm 16mm, clip]{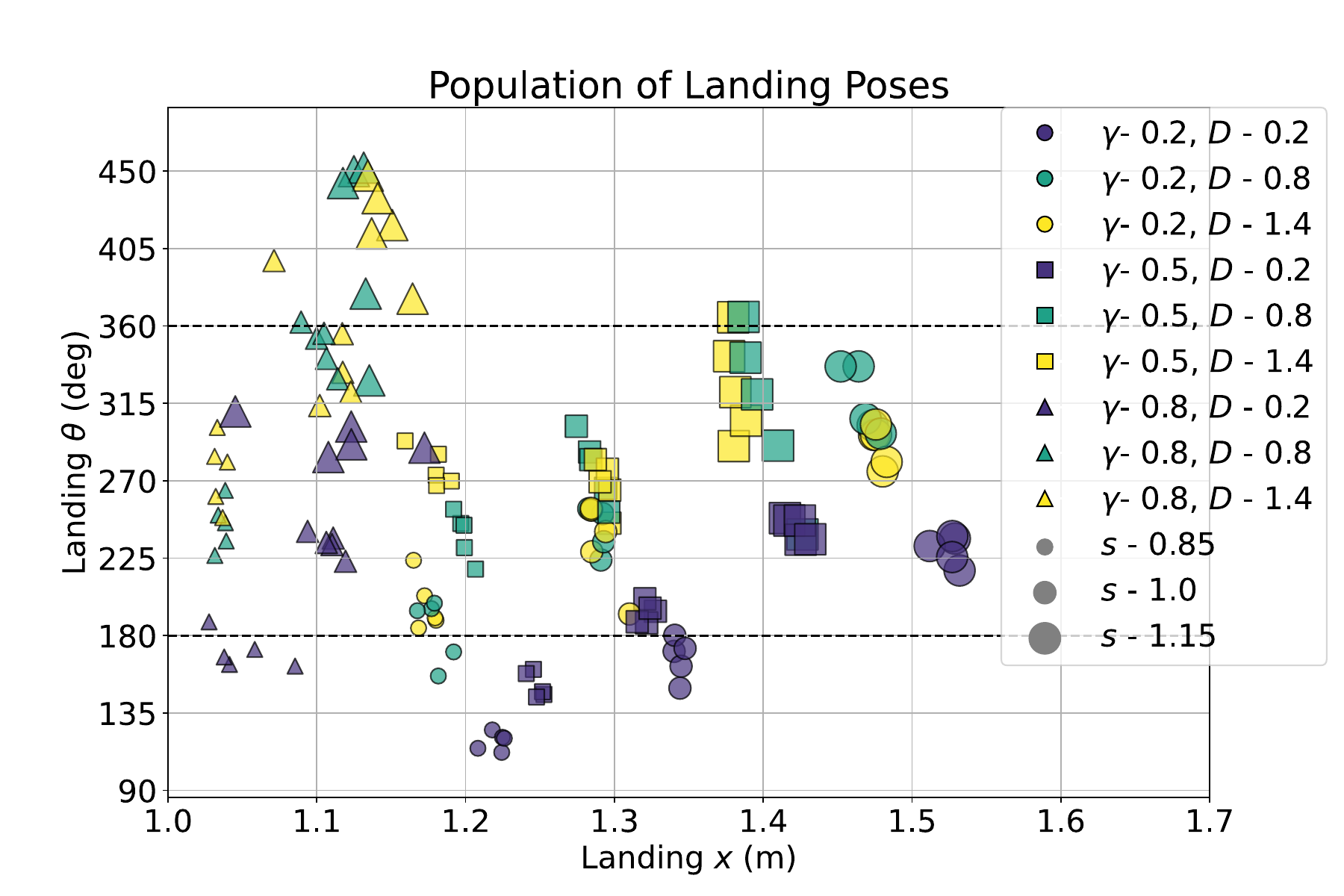}
%\includesvg[width=0.5\textwidth]{imgs/landing_x_angle_scatter}
\caption{\small Initial population of landing poses generated from impulse-momentum-based control using a mesh of $3\times3\times3=27$ commands. Each command is executed 5 times, yielding 135 throws.}
\label{fig:scatter-landing}
\end{figure}

We first verify that our impulse-momentum-based control design can yield a large set of landing poses using only 3 control parameters (\textcolor{MildRed}{pitch'' $\gamma$}, \textcolor{MildGreen}{speed'' $s$}, \textcolor{MildBlue}{``damping'' $D$}), ensuring that this family of throwing motions includes commands capable of flipping the bar with large rotations.

To validate the control design, we conducted throwing experiments for a grid of $3^3=27$ control commands (3 \textcolor{MildRed}{pitch'' $\gamma$}, 3 \textcolor{MildGreen}{speed'' $s$}, 3 \textcolor{MildBlue}{``damping'' $D$}). Each command is executed 5 times to account for intrinsic randomness in the flipping process, arising from tiny variations in the grasp pose, stochastic friction between the finger and the bar~\cite{liu2023beyond}, and the unrepeatable motor control of the Panda robot. A metal payload of 150 gram is configured in the middle of the bar, making its CoM be 12 cm from the grasp point. The resulting population of $27\times5=135$ landing poses is illustrated in Fig.~\ref{fig:scatter-landing}.
%In Fig.~\ref{fig:scatter-landing}, the three control parameters are encoded into three visual channels: shape - pitch'' $\gamma$, size - speed'' $s$, and color - ``damping'' $D$, to provide an intuitive understanding of the effect of the control parameters. 
By focusing on one parameter/channel at a time, we can observe the effect of individual control parameters:

% \begin{itemize}
%     \item \tikz \draw[fill=gray] (0,0) circle (3pt); Circle (blue)
%     \item \tikz \draw[fill=gray] (0,0) rectangle (0.25,0.25); Square (green)
%     \item \tikz \draw[fill=gray] (0.125,0.25) -- (0.25,0) -- (0.0,0.0) -- cycle; Upward Triangle (blue)
% \end{itemize}

% Define custom colors based on viridis colormap
% Define custom colors based on the provided RGBA values
\definecolor{dampingLow}{RGB}{70, 49, 126}   % (0.275191, 0.194905, 0.496005) converted to 0-255 scale
\definecolor{dampingMid}{RGB}{31, 161, 135}  % (0.122312, 0.633153, 0.530398) converted to 0-255 scale
\definecolor{dampingHigh}{RGB}{253, 231, 36} % (0.993248, 0.906157, 0.143936) converted to 0-255 scale

\begin{itemize}
    \item \textbf{``pitch'' $\gamma$:} \tikz \draw[fill=gray] (0,0) circle (3pt); $\rightarrow$ \tikz \draw[fill=gray] (0,0) rectangle (0.25,0.25); $\rightarrow$ \tikz \draw[fill=gray] (0.125,0.25) -- (0.25,0) -- (0.0,0.0) -- cycle;, increasing ``pitch'' steers the robot from a ``forward'' throw to an ``upward'' throw, resulting in a decreased landing distance, and an increased landing angle. 
     \item \textbf{``speed'' $s$:} \tikz \draw[fill=gray] (0,0) circle (2pt); $\rightarrow$ \tikz \draw[fill=gray] (0,0) circle (3pt); $\rightarrow$ \tikz \draw[fill=gray] (0,0) circle (4pt);, increasing the ``speed'' raises the linear velocity and also causes a parasitic increase in angular velocity, making the landing distance and landing angle increase simultaneously. 
    \item \textbf{``damping'' $D$:} (\tikz \draw[fill=dampingLow] (0,0) circle (4pt);, \tikz \draw[fill=dampingLow] (0,0) rectangle (0.35,0.35);, \tikz \draw[fill=dampingLow] (0.15,0.3) -- (0.3,0) -- (0.0,0.0) -- cycle;) \textcolor{dampingLow}{purple landing poses} obtained with the lowest damping are sparse and cannot achieve a full flip (360 degree), highlighting the limitations of purely velocity-based control. Introducing ``damping'' significantly expands the set of feasible landing poses. \tikz \draw[fill=dampingLow] (0,0) circle (4pt); $\rightarrow$ \tikz \draw[fill=dampingMid] (0,0) circle (4pt); $\rightarrow$ \tikz \draw[fill=dampingHigh] (0,0) circle (4pt);, increasing damping effectively increases the landing angle while slightly reducing landing distance.
\end{itemize}

Overall, our impulse-momentum-based control design significantly expands the reachable space of landing poses compared to the conventional velocity-based control design, notably enabling full 360-degree flips.
% Despite the global trends, the quantitative relationship is non-linear, as evidenced by the non-uniform effect of damping on horizontal landing distance. This implies that a global linear function to describe the command-outcome relationship is insufficient, and multiple trials may be necessary to learn the proper parameters for throwing-flipping the bar into the target box.

\subsection{Learning to throw-flip: model comparison}
% \begin{figure}[h!]
%     \centering
%     \includesvg[width=0.48\textwidth]{imgs/support_target_poses}
%     \caption{\small Mean landing poses for the four commands in the support simplex and the four unseen target poses.}
%     \label{fig:support-target}
% \end{figure}
\begin{figure}[h!]
    \centering
    \includegraphics[width=0.48\textwidth]{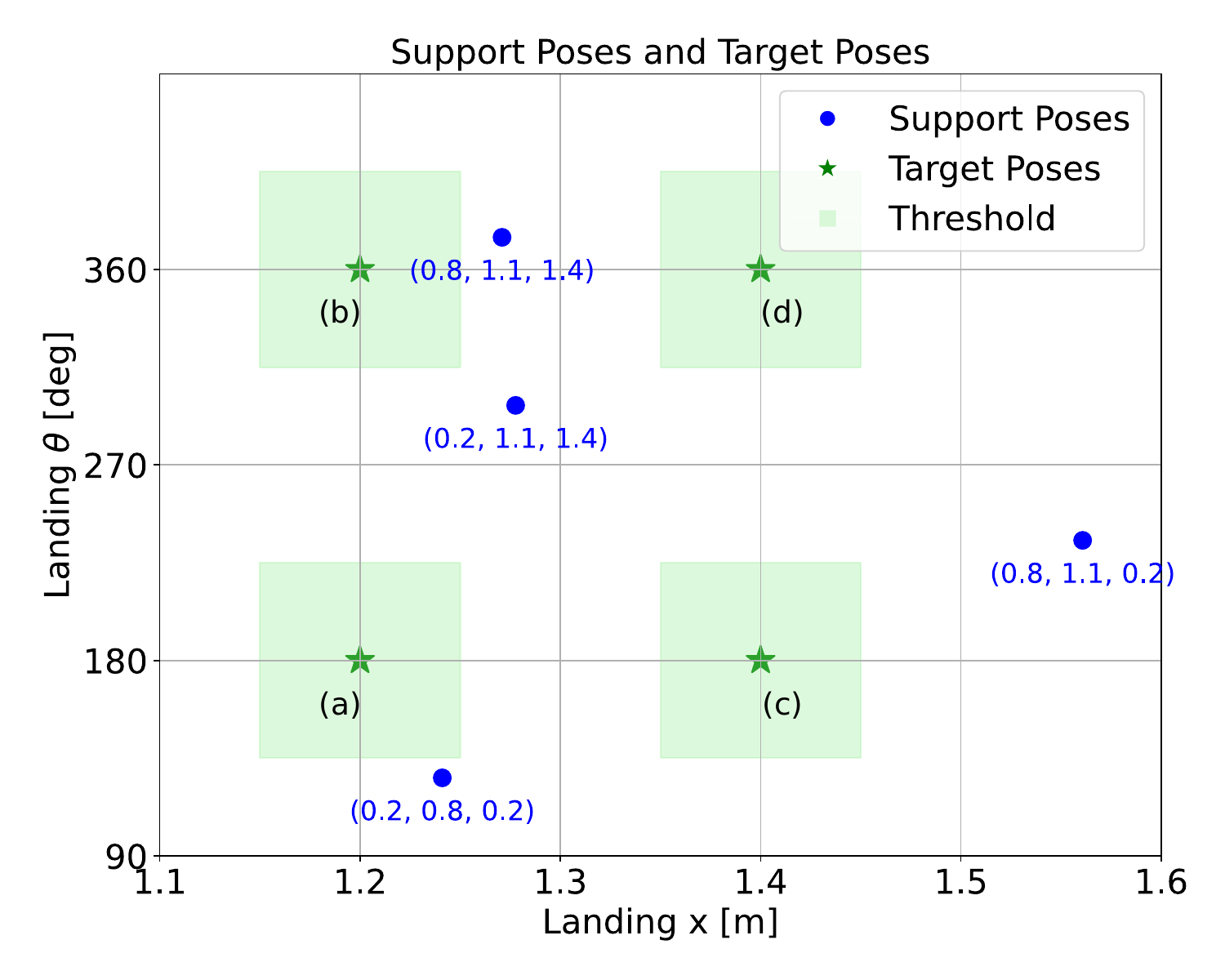}
    \caption{\small Mean landing poses for the four commands in the support simplex and the four unseen target poses.}
    \label{fig:support-target}
\end{figure}
\begin{figure*}[t!]
\centering
\includegraphics[width=\textwidth]{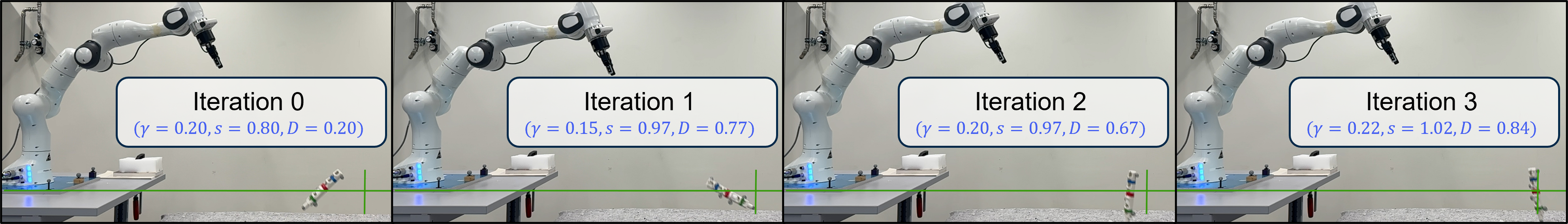}
\caption{\small A sample learning process for target pose (1.4, 180) with Model 2. Iterative 0 represents the closest landing pose within the support set. The green vertical bar represents the desired landing pose.}
\label{fig:iter_learning}
\end{figure*}
\begin{figure*}[th!]
    \centering
    \begin{subfigure}[b]{0.245\textwidth}
        % \includesvg[width=\textwidth]{imgs/iterative/progress_summary_2}
        \includegraphics[width=\textwidth]{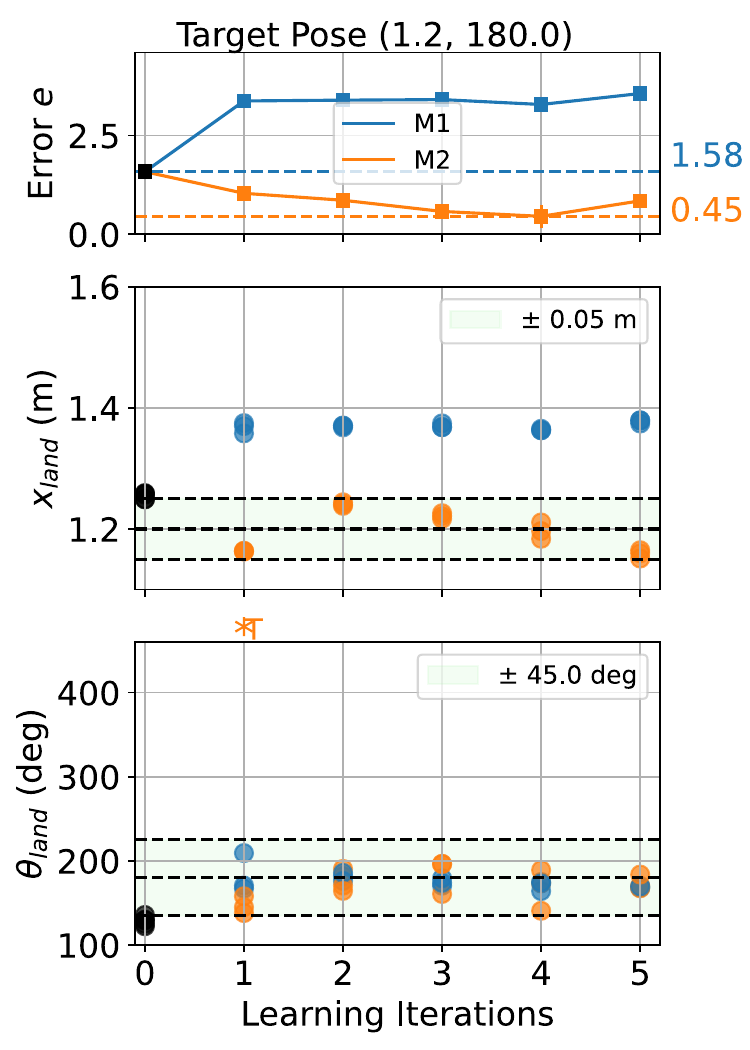}

        \caption{\small}
        \label{fig:iter_target3}
    \end{subfigure}
    \begin{subfigure}[b]{0.245\textwidth}
        % \includesvg[width=\textwidth]{imgs/iterative/progress_summary_0}
        \includegraphics[width=\textwidth]{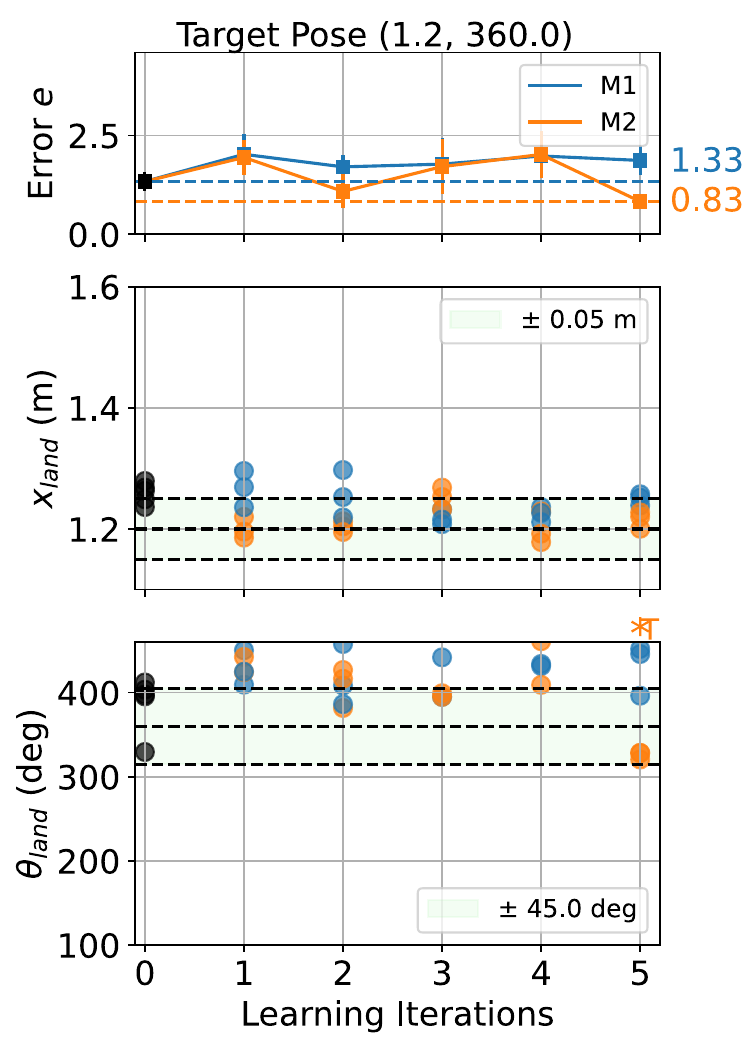}
        \caption{\small }
        \label{fig:iter_target1}
    \end{subfigure}
    \begin{subfigure}[b]{0.245\textwidth}
        % \includesvg[width=\textwidth]{imgs/iterative/progress_summary_3}
        \includegraphics[width=\textwidth]{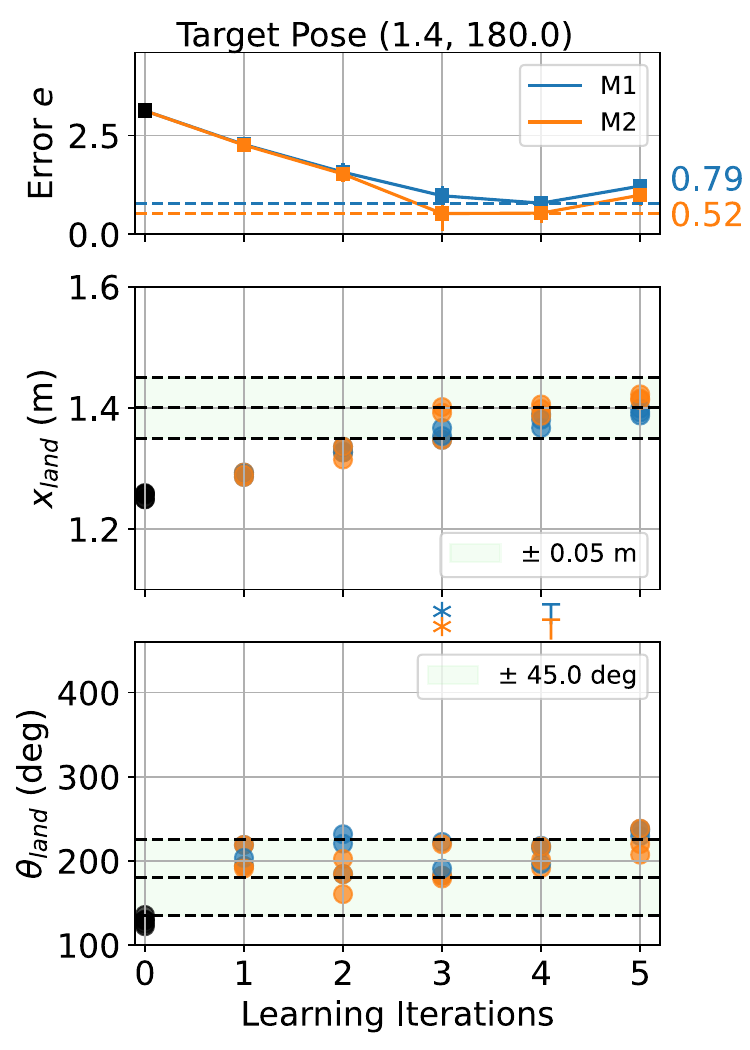}
        \caption{\small }
        \label{fig:iter_target4}
    \end{subfigure}
    \begin{subfigure}[b]{0.245\textwidth}
        % \includesvg[width=\textwidth]{imgs/iterative/progress_summary_1}
        \includegraphics[width=\textwidth]{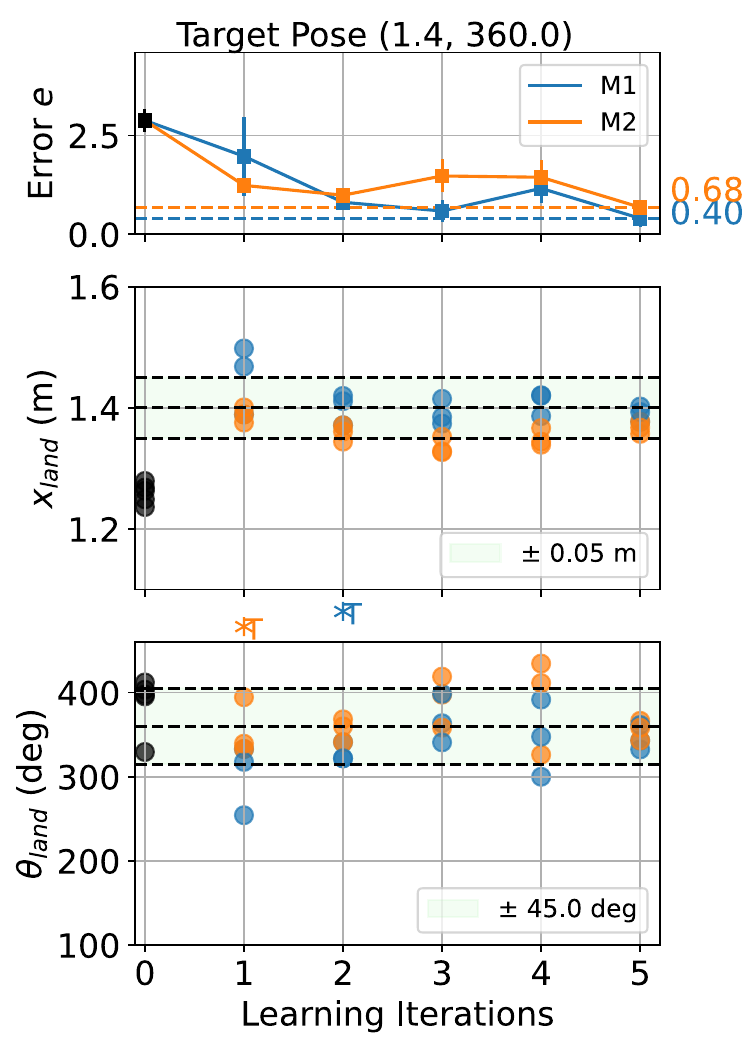}
        \caption{\small }
        \label{fig:iter_target2}
    \end{subfigure}
    \caption{Summary of iterative learning of four unseen target poses. Symbol `*' represents the first iteration where the landing pose falls within the position and orientation threshold for \textbf{at least 2 out of 3 trials}. Symbol `T' represents the first iteration where the landing pose falls within the position and orientation threshold for \textbf{all the 3 trials}.}
\label{fig:iter_summary}
\end{figure*}

We report on a comparative analysis of throw-flip performance using end-to-end learning (model 1) as opposed to projectile-dynamics-based learning (model 2). Given the inherent stochasticity in the throwing process, as demonstrated by the variance of landing poses in Fig.~\ref{fig:scatter-landing}, we set the target tolerance to ($\pm$5 cm, $\pm$45 degrees). To evaluate the capability of the learning and adaptation system in throw-flipping to unseen poses, we provide only four initial commands that form a simplex in the 3D command space, offering a minimal non-degenerate representation of the forward model. Four unseen target landing poses are specified: (a) (1.2, 180), (b) (1.2, 360), (c) (1.4, 180), (d) (1.4, 360). The mean landing poses associated with the four commands in the support simplex, along with the target poses and their tolerances, are illustrated in Fig.~\ref{fig:support-target}. To account for the stochasticity in the throwing process, each command is executed three times per iteration to obtain the mean outcome. Each learning path is conducted for five iterations, resulting in 15 throwing trials per learning path. Results are summarized in Table~\ref{tab:iter_table} and Fig.~\ref{fig:iter_summary}.  A sample learning process for the target (1.4, 180) with Model 2 is illustrated in Fig.~\ref{fig:iter_learning}. 

% \begin{figure}[ht!]
%     \centering
%     \includesvg[width=0.5\textwidth]{imgs/simplex_commands}
%     \caption{\small The simplex of commands used as the support set for learning to throw-flip.}
%     \label{fig:support-simplex}
% \end{figure}

\begin{table}[h!]
\centering
\begin{tabular}{c|ccccc|cc}
\hline
{} & Error & \multicolumn{2}{c}{First Iter.} & \multicolumn{2}{c}{Min.} & \multicolumn{2}{|c}{$\#$Iters Enter Thres.}\\
\cmidrule(lr){3-4} \cmidrule(lr){5-6} \cmidrule(lr){7-8}
 {Target} & Initial & M1 & M2 & M1 & M2 & M1 & M2\\ 
\hline
(a) & 1.58 & 3.37 & \textbf{1.04} & 1.58 & \textbf{0.45} & Failed & \textbf{1} \\ %\hline
(b) & 1.33 & 2.02 & \textbf{1.94} & 1.33 & \textbf{0.83}  & Failed & \textbf{5} \\ %\hline
(c) & 3.12 & 2.27 & \textbf{2.25} & 0.79 & \textbf{0.52} & 4 & 4 \\ %\hline
(d) & 2.88 & 1.97 & \textbf{1.24} & \textbf{0.40} & 0.68 & 2 & \textbf{1} \\ \hline
Average & 2.23 & 2.41 & \textbf{1.62} & 1.02 & \textbf{0.62} & >4.5 & \textbf{2.75}  \\ \hline
\end{tabular}
\caption{Summary of normalized error and the number of iterations required to reach the target threshold for the four target poses. `M1' - Model 1, `M2' - Model 2.}
\label{tab:iter_table}
\end{table}
\captionsetup[subfigure]{skip=2pt}  
% \noindent \textbf{Discussion.} We discuss the results in the following aspects:\\
\noindent \textbf{Normalized errors:} %although the iterative online learning procedure is path-dependent—earlier iterations influence the subsequent learning path due to the differences in the data gathered along the way—the nearest neighbors selected at the start are identical for Model 1 and Model 2. This enables a direct comparison of accuracy between the two forward models. 
The normalized error for Model 2 after the first iteration is consistently lower than that of Model 1, indicating that incorporating physical knowledge of nonlinear projectile dynamics explains away part of the nonlinearity in the command-outcome map. On average, Model 2 achieves 40\% smaller error than Model 1.\\
\noindent \textbf{Iterative improvement:} For Model 2, the iterative learning procedure progressively brings the landing pose closer to the target. This is evidenced by the fact that the minimum error is always achieved after the second iteration, despite some error oscillation.\\
\noindent \textbf{First iteration entering threshold:} Model 2, which incorporates projectile dynamics, reaches the target threshold in fewer iterations than Model 1 for all four target poses.\\
\noindent \textbf{Error correction:} Our system design demonstrates the ability to escape local minima. In particular, for the fourth iteration of (Target (b), Model 2) and the fourth iteration of (Target (d), Model 1), the landing pose exhibits greater error than in previous iterations, indicating that the neighbor-constructed local model was inaccurate. However, a new set of nearest neighbors is selected thanks to the re-selection routine, allowing the system to correct the large error in the next iteration.\\   
% \noindent \textbf{Local minima:} Despite the improvements from incorporating physics-based modeling and an error-correction routine, the system can still suffer from local minima. For Target (a) and (b), Model 1 failed to find valid throwing commands after five iterations.

% \noindent \textbf{Remark on Statistical Evaluation:}\\
% Due to the manual nature of the experimental procedure, we prioritized evaluating performance across a diverse set of unseen target poses rather than repeating multiple learning paths for each target. As a result, only one learning trajectory was executed per target pose. To mitigate stochasticity in the landing outcome, each command was executed three times per iteration, and the mean landing pose was used. We report the metric “number of trials entering the threshold” (an $\ell_\infty$-style criterion), which reflects task success more directly than mean or standard deviation of landing error. Given the small number of trials per command, computing standard deviations would be statistically weak and potentially uninformative. We acknowledge the limited scale of experimentation and leave large-sample statistical evaluation and automation of repeated trials to future work.

\subsection{Transfer learning under CoM shift}
In this experiment, we demonstrate \textbf{Model 3: transfer learning} can accelerate iterative learning for new objects under CoM shift. The transfer learning method follows the procedure described in Section III.G. Compared to the previously thrown bar, we shift the metal payload from the middle to the furthest position away from the tip, effectively changing the object's CoM offset from 12 cm to 18 cm.

\begin{figure}[h!]
    \centering
\includegraphics[width=0.48\textwidth]{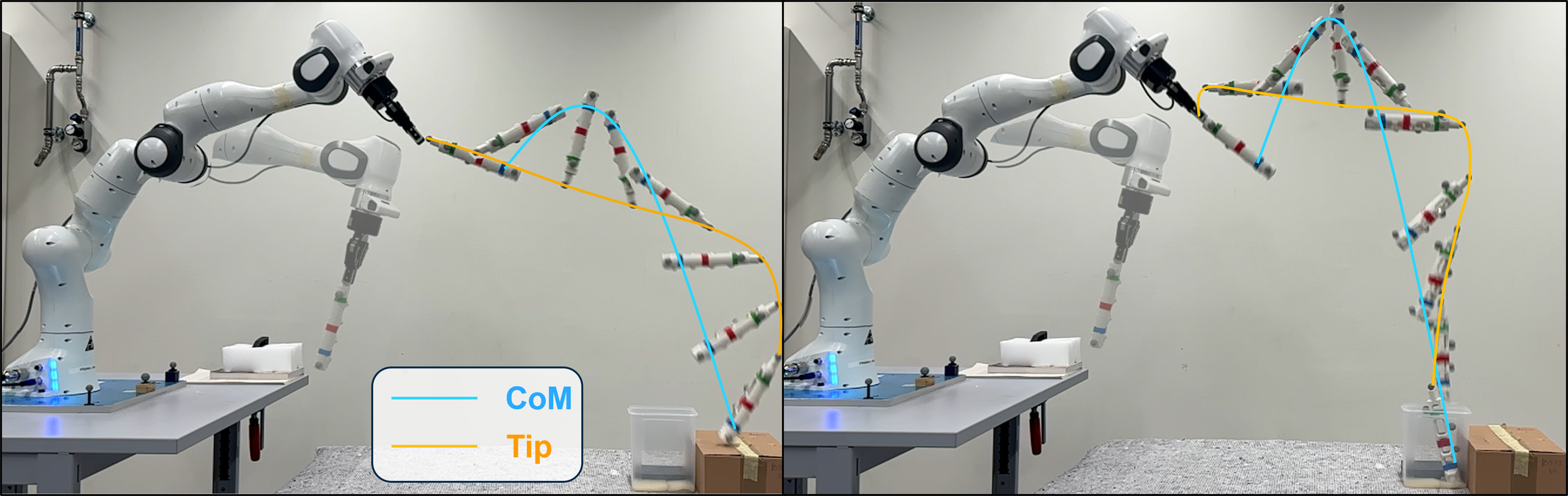}
    \caption{\small Transfer learning under CoM shift. Left: the command to throw-flip the original bar is invalid for the CoM-shifted bar. Right: A new valid command is found after transfer learning within few iterations.}
    \label{fig:inbox_transfer}
\end{figure}

\begin{figure}[t!]
    \centering
\includegraphics[width=0.5\textwidth]{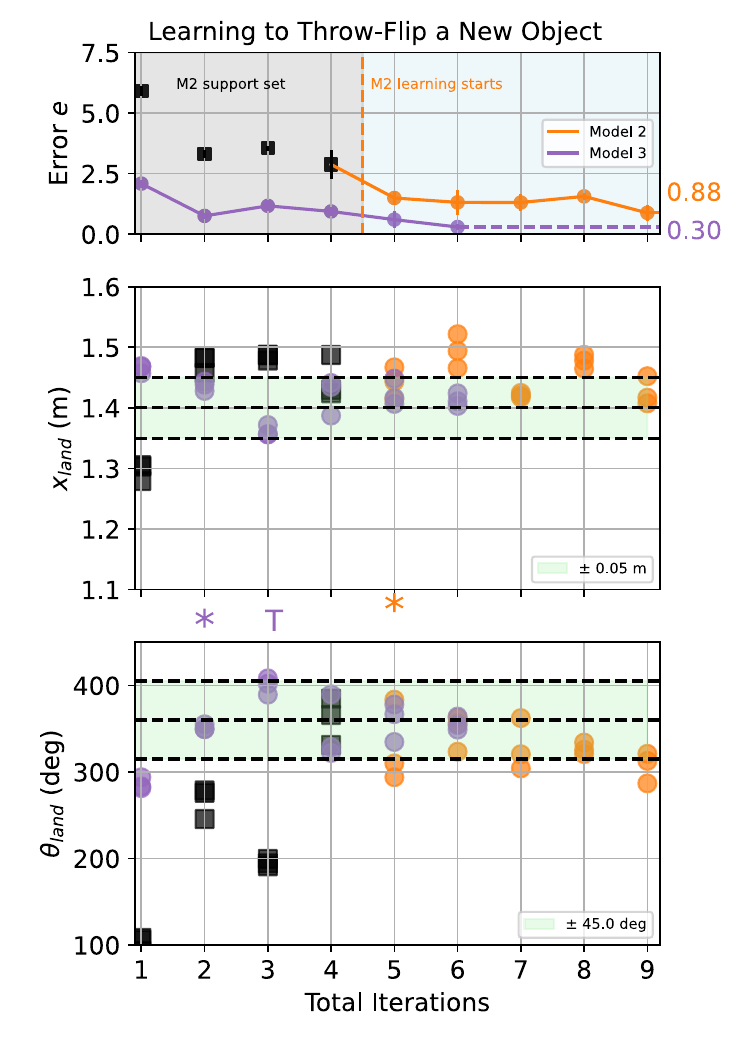}

    % \includesvg[width=0.5\textwidth]{imgs/iterative/transfer_summary_0}
    \caption{\small Summary of transfer learning. Symbol `*' represents the first iteration where the landing pose falls within the position and orientation threshold for \textbf{at least 2 out of 3 trials}. Symbol `T' represents the first iteration where the landing pose falls within the position and orientation threshold for \textbf{all the 3 trials}.}
    \label{fig:transfer_summary}
\end{figure}

% \begin{figure}[h!]
% \centering
% \includegraphics[width=0.48\textwidth]{imgs/iterative/iter_transfer_new.png}
% \caption{\small Progress of transfer learning with Model 3.}
% \label{fig:iter_learning}
% \end{figure}

Fig.~\ref{fig:inbox_transfer} illustrates the effect of increased landing position under CoM shift and the result of transfer learning. The learning progress is summarized in Fig.~\ref{fig:transfer_summary}. For comparison, we also include a baseline approach that does not incorporate physical knowledge of CoM shift. In this baseline, the support set simplex is gathered using four commands, followed by standard iterative learning with Model 2. \\
% For simplicity, we refer to this benchmark procedure as ``Model 2''. \textbf{Discussion:}\\
\noindent \textbf{First iteration entering threshold:} 
Strikingly, transfer learning with Model 3 reaches a 2/3 success rate after two iterations and 3/3 success rate after three iterations, even before entering the normal iterative learning process. In contrast, Model 2 requires a total of 5 iterations to reach 2/3 success rate, and failed to achieve 3/3 success rate after 9 iterations. This demonstrates Model 3's ability to effectively \emph{reuse} past experience when throwing unseen objects, leveraging physical knowledge for faster adaptation.
\noindent \textbf{Iterative improvement:} 
After 6 iterations, Model 3 achieves a significantly lower normalized error than Model 2 (0.3 vs. 0.88). This improvement is likely due to the fact that the data population gathered by Model 3 is much closer to the target (as indicated by the smaller normalized errors in the first three iterations) compared to the standard support simplex used in Model 2. Consequently, the local approximation quality in Model 3 remains higher throughout learning, leading to better decisions during the iterative process.

%\subsection{Human %comparison}
%\textcolor{blue}{TODO (Elise): compare adaptation capability with humans, demonstrating superhuman adaptation performance}
\section{Conclusion}
In this work, we introduce a learning and adaptation framework that enables a robot to throw-flip objects with desired landing positions and orientations. By leveraging the impulse-momentum principle, we formulate a minimal three-parameter control system that effectively decouples the effects of the parasitic rotation in revolute robots, significantly expanding the set of feasible landing poses. 

To address the challenge posed by the absence of accurate forward models in dynamic manipulation, we propose a model-based learning framework that seamlessly integrates data-driven adaptation with physical knowledge. Our experiments demonstrate that:
\begin{itemize}
    \item Incorporating projectile dynamics improves learning efficiency by reducing sample complexity.
    \item Data on object in-hand spinning motion can be reused to accelerate transfer learning for new objects with Center of Mass (CoM) shifts, reducing the need to gather a support set from scratch.
\end{itemize}
As a result, our learning system can throw-flip a new object within dozens of trials, requiring only a small support set of four commands and achieving fast adaptation within five iterations. This contrasts strongly with end-to-end learning methods, which require thousands of throwing experiments to achieve the desired landing pose. Although CoM information is needed for fast learning, the object's CoM can be easily obtained, e.g. through a string hanging experiment or by using F/T sensor measurements when the robot holds the object in different poses~\cite{gaz2017payload}. 

One limitation of the current control system is the lack of repeatability in throwing outcomes when executing the same robot-throwing trajectory, necessitating multiple trials per command. In the future, it would be interesting to explore the learning of robust throwing motions to reduce landing variance, improving the consistency and reliability of throw-flipping. It is also interesting to explore data assimilation with richer sensing modalities, e.g. wrench sensing.
% \input{sections/04_experiments}
% \input{sections/05_discussion}
% \input{sections/06_conclusion}


% \section*{ACKNOWLEDGMENT}
% \noindent The authors acknowledge the support of the H2020 EU project DARKO under Grant Agreement No. 101017274.

\bibliographystyle{IEEEtran}
\bibliography{references.bib}

\end{document}